\pdfoutput=1

\documentclass[11pt]{article}

\usepackage{EMNLP2023}

\usepackage{times}
\usepackage{latexsym}

\usepackage[T1]{fontenc}

\usepackage[utf8]{inputenc}

\usepackage{microtype}

\usepackage{inconsolata}

\usepackage[normalem]{ulem}
\usepackage{graphicx}
\usepackage{booktabs}
\usepackage{wrapfig}
\usepackage{multirow}

\definecolor{mygreen}{HTML}{2CB600}
\usepackage{xcolor}

\newcommand\extrafootertext[1]{%
    \bgroup
    \renewcommand\thefootnote{\fnsymbol{footnote}}%
    \renewcommand\thempfootnote{\fnsymbol{mpfootnote}}%
    \footnotetext[0]{#1}%
    \egroup
}

\usepackage{algorithm}
\usepackage[noend]{algcompatible}
\newcommand{\algorithmicinput}{\textbf{input}}
\newcommand{\INPUT}{\item[\algorithmicinput]}
\newcommand{\algorithmicoutput}{\textbf{output}}
\newcommand{\OUTPUT}{\item[\algorithmicoutput]}

\newcommand{\algorithmiccode}{\textbf{execute}}
\newcommand{\CODE}{\item[\algorithmiccode]}

\newcommand{\comment}[3]{{\small{\textcolor{#3}{[#1 #2]}}}}
\renewcommand{\comment}[3]{}  

\newcommand{\com}[1]{}
\newcommand{\roee}[1]{\comment{\marker{ROEE}}{#1}{red}}
\newcommand{\enav}[1]{\comment{\marker{ENAV}}{#1}{purple}}

\newcommand{\yoad}[1]{\comment{\marker{YOAD}}{#1}{orange}}
\newcommand{\yonatan}[1]{\comment{\marker{YONATAN}}{#1}{blue}}

\newcommand{\resolved}[1]{}

\title{q2d: Turning Questions into Dialogs to Teach Models How to Search}

  \newcommand{\authorspace}[0]{\quad}

\author{
  Yonatan Bitton$^{\dagger,\ddagger,*}$ \authorspace 
 Shlomi Cohen-Ganor$^{\ddagger}$
 \authorspace
 Ido Hakimi$^{\ddagger}$\authorspace 
 \\
 \textbf{Yoad Lewenberg}$^{\ddagger}$\authorspace 
 \textbf{Roee Aharoni}$^{\ddagger}$\authorspace 
 \textbf{Enav Weinreb}$^{\ddagger}$\authorspace \\
 $^{\dagger}$ The Hebrew University of Jerusalem; 
$^{\ddagger}$ Google Research \\
  yonatan.bitton@mail.huji.ac.il \\  
  \authorspace \{yonatanbitton,shlomic,idohakimi,yoadlew,roeeaharoni,eweinreb\}@google.com \\
}

\begin{document}
\maketitle

\extrafootertext{$^{\ddagger}$Work done during an internship at Google Research.}

\begin{abstract}
One of the exciting capabilities of recent language models for dialog is their ability to independently search for relevant information to ground a given dialog response. However, obtaining training data to teach models how to issue search queries is time and resource consuming.
In this work, we propose \emph{q2d}: an automatic data generation pipeline that generates information-seeking dialogs from questions. We prompt a large language model (PaLM) to create conversational versions of question answering datasets, and use it to improve query generation models that communicate with external search APIs to ground dialog responses. Unlike previous approaches which relied on human written dialogs with search queries, our method allows to automatically generate query-based grounded dialogs with better control and scale.
Our experiments demonstrate that: (1) For query generation on the QReCC dataset, models trained on our synthetically-generated data achieve 90\%--97\% of the performance of models trained on the human-generated data; (2) We can successfully generate data for training dialog models in new domains without any existing dialog data as demonstrated on the multi-hop MuSiQue and Bamboogle QA datasets. (3) We perform a thorough analysis of the generated dialogs showing that humans find them of high quality and struggle to distinguish them from human-written dialogs.

\end{abstract}

\section{Introduction}
\begin{figure}[!tb]    \includegraphics[width=\columnwidth]{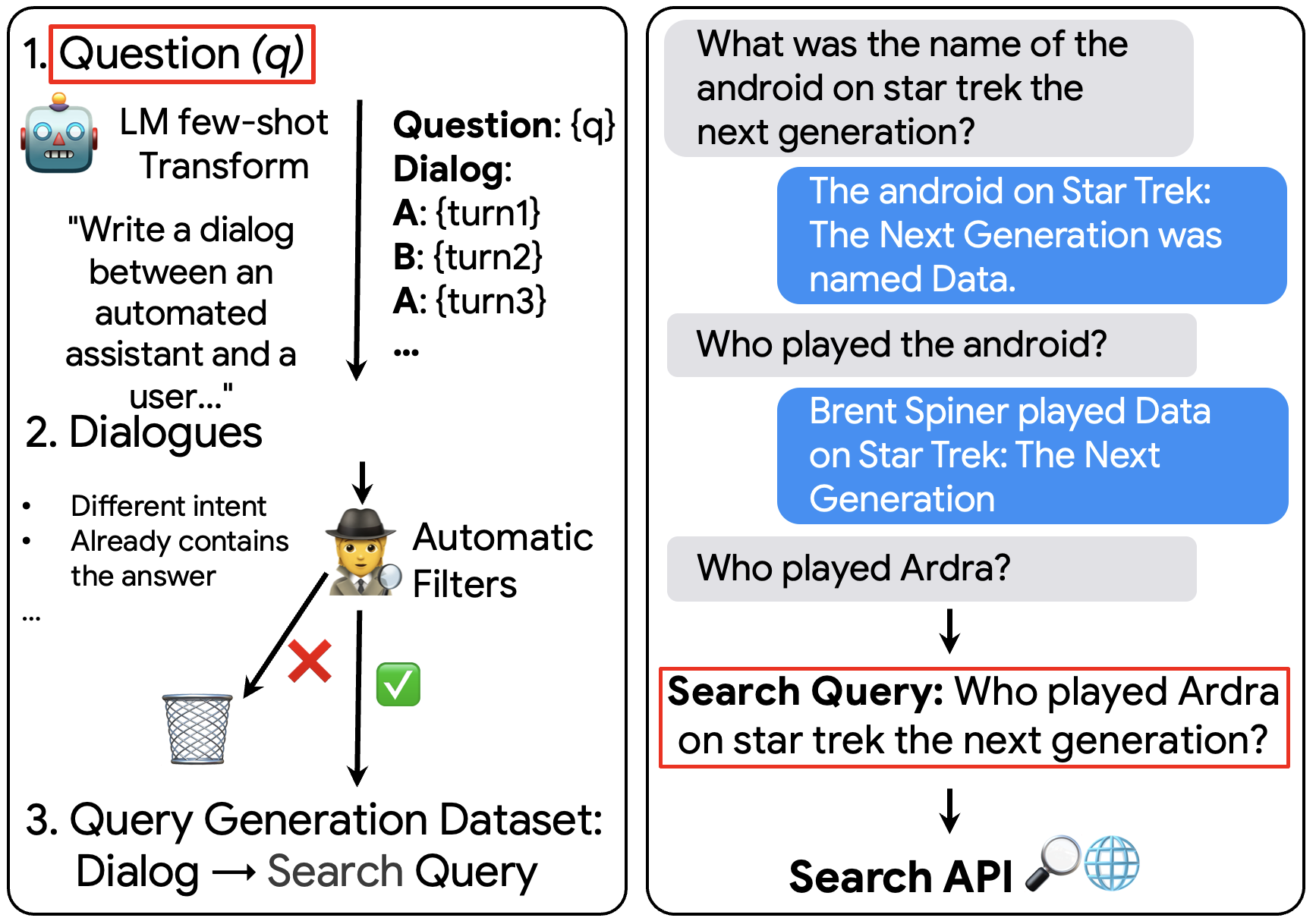}\\
    \caption{Left: Our \emph{q2d} method starts from an existing query or question and prompt a few-shot language model to transform it into a dialog. We filter out cases where the intent of the generated dialogue differs from the intent of the initial query and apply additional filters. \\
    Right: We take a question from the QReCC dataset (surrounded by a rectangle) and generate an information-seeking dialog with \emph{q2d}. By starting with a query and generating a dialog, we create \{dialogue $\rightarrow$ query\} dataset, which is used to train and evaluate query generation models, which communicate with an external search API to generate factual responses. \roee{I think it will be clearer to show a concrete example that goes from a concrete search query -> to PaLM -> to a concrete generated dialog. }\yonatan{It was as you said before - but Amir Golberson it's better to "reverse" it, showing first the dialog, then the search query, in order to understand the "final" usage of the query generation model. Can you think about this again in this context?}
    }
    \label{fig:fig1}
\end{figure}

Recent dialog generation models, such as LaMDA \cite{thoppilan2022lamda}, 
BlenderBot3 \cite{shuster2022blenderbot} and Sparrow \cite{glaese2022improving} use an external search API to generate grounded and factually accurate responses \cite{parisi2022talm}. This is important for providing reliable and consistent answers \cite{shuster2022language}, especially when discussing entities and asking related questions with anaphora. To do this, these models use a query generation component that is trained on dialog-to-search-query datasets. When the model is triggered with a dialog turn that requires search, it generates a query that is used to obtain a search result, which is then used to generate a grounded response. This allows the model to provide relevant information about the world in its responses to user queries. For example, a model trained in 2021 should be able to provide a factual response to the question ``How old is Joe Biden?'' even in 2023. In a conversation, one might discuss an entity (e.g. ``Joe Biden'') and later ask a related question (e.g. ``How old is he?'') with anaphora. In order to provide reliable and consistent answers, it is necessary to generate a decontextualized query (e.g., ``How old is Joe Biden'') for a search engine.

Using APIs also decouples language and reasoning from knowledge \cite{borgeaud2021improving,parisi2022talm}, which can help prevent errors caused by outdated information being stored in the model's parameters. For example, if a model trained at the end of 2021 is asked ``How old is the current president?'', it may produce the incorrect query ``How old is Donald Trump'' if its parameters are outdated or if it provides factually-inconsistent responses (a.k.a ``hallucinations'').

Query generation datasets have been created using human annotators, limiting them in scale, control, and quality \cite{komeili2021internet}. As a result, when a new domain is introduced, a significant amount of human effort is required to create a new query generation dataset for that domain \cite{gupta2021dialfact,dziri2021evaluating}. The fact that language models often generate hallucinations \cite{zhao2020reducing,maynez2020faithfulness,lee2018hallucinations}, especially in new domains or dialogs that differ from the training data \cite{nie2020like,honovich2021q, honovich2022true}, highlights the need for more effective query generation datasets that will foster more grounded and factually consistent models.

In this work, we propose a data generation pipeline to improve grounded dialog models with access to search engines. To create a dialog-to-search-queries dataset for training the query generation component in such models, we reverse the process, starting from a search query and generating an information-seeking dialog that corresponds to that query. Our automatic pipeline, shown in Figure~\ref{fig:fig1}, begins with a search query or question, and prompts a large language model \cite[PaLM;][]{chowdhery2022palm} to generate a conversational dialog that conveys the information need implied by the given query. For example in Figure~\ref{fig:fig1}, we take the question ``Who played Ardra on star trek the next generation?'' from the Natural Questions dataset \cite{kwiatkowski2019natural} and generate a dialog with a similar intent: the correct answer to the original question (``Marta DuBois'') is also a correct response to the generated dialog. This process allows us to leverage existing question-answering datasets, which are widely available for different domains, and extend them by generating dialogs that preserve the original information need while controlling the dialog domain and style. 

To assess whether the automatically generated dialogs can replace human-generated dialogs, we experiment with QReCC NQ \cite{anantha2020open}, a human-curated dialog dataset. We generate a training set that is the same size as the original dataset, but with synthetic dialogue, and use it to train a query generation model. The resulting model obtains 90\%--95\% of the performance of models trained on the human-generated training data, using the same metrics used to evaluate QReCC \cite{anantha2020open}. 

Other than training query generation models, our approach is also useful for training the dialog generation models themselves when no dialog data is available for a new domain. We demonstrate that on the domain of multi-hop question answering , where we first show that existing dialog models struggle to perform well on a domain-specific challenge set. We then generate synthetic dialog data from the MuSiQue \cite{trivedi2021musique} multi-hop QA dataset, and show that training a dialog model on this data improves performance. 

We provide a thorough analysis of the quality of the generated datasets, demonstrating that they (a) looks natural, and humans struggle to distinguish the synthetic dialogs from natural; (b) factual: generated and human-annotated answers perform similarly in query generation; (c) correct: dataset labels are accurate, and strict filtering improves results.

To conclude, our main contributions are: 
\begin{enumerate}
    \item We introduce \emph{q2d}: an automatic method to generate information-seeking dialogs from questions using large language models.
    \item We show that our method is beneficial for training query generation and dialog generation, including in different domains like multi-hop QA.
    \item A thorough analysis showing that the synthetically generated dialogs are natural, factual and correct. 
\end{enumerate}


\section{Generating Dialogs from Questions}
\label{sec:generation}
\begin{algorithm}[!t]
  \caption{Generate Dialogues from Questions}
  \label{DolphinnPrep}
  \begin{algorithmic}
    \INPUT{\\Few-Shot Model $M_{fs}$, QA Dataset $(Q,A)$,\\ Examples Queries $S_q=\left\{\left(q_{i}, d_{i}\right)\right\}_{i=1}^{k}$, \\ 
    Examples Dialogues $S_d=\left\{\left(d_{i}, q_{i}\right)\right\}_{i=1}^{k}$, \\ Instructions Query $I$, Instructions Dialogue $I_r$, }
    \CODE{
    \State $dataset \gets \emptyset$
    \FOR{$(q,a) \in (Q,A)$} 
      \State $dialogue \gets M(S_q, I, q)$ 
          \State $q' \gets M_{fs}(S_d, I_r, dialogue)$
        \IF{filter(dialogue,q,q',a)}
            \State
          $dataset.add((dialogue,q,a))$ 
        \ENDIF
    \ENDFOR
    }
  \OUTPUT{\\Query Generation Dataset: $D=\left\{\left(d_{i}, q_{i}\right)\right\}_{i=1}^{|Q|
  }$}
  \end{algorithmic}
  \label{algorithm}
\end{algorithm}

In this section, we describe our automatic method, called \emph{q2d}, for generating dialogs from questions, and the properties of datasets produced by this method. Our goal is to reduce the effort associated with generating a training dataset for training generation, and to improve query-generation-based dialog models with a high-quality training dataset. Query generation can start by extracting queries from existing dialogs. However, our approach is unique in that it begins with factual queries or questions, allowing us to leverage existing resources. Any question-answering dataset, queries dataset, or queries used in popular web search services or dialog model logs can be used with our algorithm. 

The algorithm is described in Algorithm~\ref{algorithm} and consists of three main steps:
\begin{enumerate}
    \item Starting from a query or question from the set $Q$, we use a few-shot model $M_{fs}$, specifically we use PaLM, and instructions $I$ to generate a dialog given the query. The few-shot prompts can be manually written to adapt to different conversation styles, or sampled from existing dialogs dataset. 
    \item Using the same few-shot examples in reverse, $S_d$ and $I_r$, we generate a query based on the generated dialog, $q'$.
    \item Filtering: we filter dialogs with different intent, or dialogs where the dialog answer is contained in the dialog. We elaborate on the different filters below. 
\end{enumerate}

\paragraph{Filtering.} In this part we attempt to filter (dialog, query) samples that would not be beneficial for training or testing. We do it in three steps, elaborated below. We stress that there are many more filtering strategies possible, and exploring them is left for future work. First, we filter out dialogs whose intent is different from the original query by measuring the similarity between the query and its reversed version using SBERT similarity ($sim(q,q')$) and comparing it to a threshold ($T_{query}$). If the similarity is below the threshold, the generated query is considered to have a different intent and the dialog is filtered. Appendix~\ref{sec:appendix}, Section~\ref{sec:filtering_examples} shows several examples of dialogs, original and reversed query and SBERT semantic similarity\yonatan{Enav - is this better? I just removed the example you didn't like, and have a full table of examples in the Appendix}. Second, we filter out cases where the answer is included in the dialog by measuring the n-gram overlap between the dialog and the answer using the Rouge metric \cite{lin2004rouge}. If the overlap is above a threshold ($T_{answer}$), the answer is entailed in the dialog and the example is filtered. For example, if the final answer (``Marta DeBois'') would have been already written in the dialog for the role of playing \textit{Ardra}, the final question (``Who played Ardra'') would not make sense. Finally, we filter out cases where the last turn of the dialog is similar (>80\%) to the original question using SBERT similarity. These cases include situations where no anaphora is required.


In this work, we use PaLM \cite{chowdhery2022palm}, a large language model with 540B parameters, as the few-shot language model for generating dialogs with a temperature of $0.6$. We provide a fully working code with GPT-3 \cite{brown2020language} for reproducibility. The set of prompts and instructions can be found in Appendix~\ref{sec:appendix}, Section~\ref{sec:appendix_prompts}. For the similarity metric ($sim$), we use the \textit{all-mpnet-base-v2} model from Sentence Transformers, with a threshold similarity of $T_{query}=0.999$. This threshold is justified through human-evaluation and ablation studies for the filtering in Section~\ref{sec:correctness}.
\enav{I don't think this is the right place for this last pararaph}\yonatan{Eliminated. Is it better?}

\section{Replacing Human-Annotated with Auto-Generated Data}
\label{sec:replacing}
\begin{figure}[!tb]
    \includegraphics[width=\columnwidth]{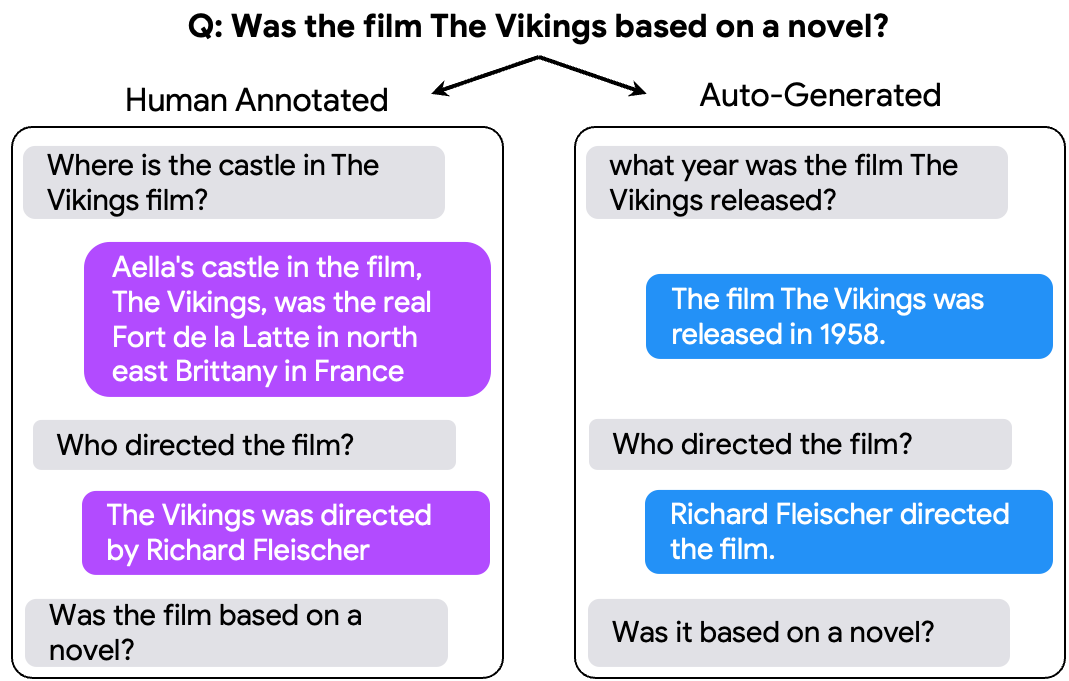}\\
    \caption{An example of human annotated dialogue from QReCC and an automatically generated dialogue produced for the same question.
    }
    \label{fig:human_vs_synthetic}
\end{figure}

In this section, we evaluate the extent to which our automatically generated dataset can replace the human-annotated dataset. We use the QReCC \cite{anantha2020open} NQ dataset, which contains (dialog, query) pairs, and automatically generate a dialog from natural questions. This allows us to create an automatically generated train set of the same size, and compare it to the human-annotated dataset. An example of a human-generated dialog compared to an automatically generated dialog is shown in Figure~\ref{fig:human_vs_synthetic}. We use the version of the dataset where the intermediate questions are contextualized, rather than decontextualized. For example, the second and third user turns in the figure are contextualized versions of the decontextualized questions ``Who directed the film, The Vikings?'' and ``Was the film The Vikings based on a novel?''.

\paragraph{Dataset Generation.} To generate our dataset, we use our \emph{q2d} method as described in Section~\ref{sec:generation}. For the few-shot examples of queries and dialogs ($S_q$ and $S_d$), we sample 15 examples from QReCC that fit within the maximum input sequence length. These examples are available in Appendix~\ref{sec:appendix}, Section~\ref{sec:appendix_prompts}. For the base questions ($Q$), we use the Natural Questions \cite{kwiatkowski2019natural} dataset instead of the QReCC NQ questions to reduce dependence on QReCC. Importantly, all of the questions and dialogs in the natural and automatically generated datasets are disjoint. In total, we generate 13K samples, the same as the QReCC NQ train set. Full prompts, instructions and examples are available in Appendix~\ref{sec:appendix}, Section~\ref{sec:examples}. 

\begin{table}[!t]
\resizebox{\columnwidth}{!}{
\begin{tabular}{@{}lllll@{}}
\toprule
Model                  & Training Dataset & \begin{tabular}[c]{@{}c@{}}SBERT \\ Similarity\end{tabular}  & \begin{tabular}[c]{@{}c@{}}Rouge-1\\ Recall\end{tabular}  & \begin{tabular}[c]{@{}c@{}}Search Results\\ Recall@10\end{tabular}  \\ \midrule
\multirow{2}{*}{T5}    & Human Annotated                   & 92.4                                                          & 88.1                                                       & 68.5                                                                 \\
                       & Auto Generated                    &
                       87.5 (\textbf{95\%})                                                          & 83.3 (\textbf{95\%})                                                       & 61.5 (\textbf{90\%})                                                                 \\ \bottomrule
\end{tabular}}
\caption{Results on the human-annotated QReCC NQ test set, experimenting with replacing the human-annotated data with automatically generated data with the \emph{q2d} method. Bold shows the percentage of performance for a model trained with auto-generated data out of a model trained with human-annotated data. Training on the automatically generated data achieves 90\%-95\% of the model trained on the human annotated results.}
\label{tab:results_replaing}
\end{table}

\paragraph{Metrics and Models.} Our metrics are the same as those used in the QReCC dataset, comparing the original and generated queries. These include Rouge-1 Recall \cite{lin2004rouge} for measuring the similarity between two text unigrams, and SBERT embedding semantic similarity for comparing the semantic content of two sentences (same metric as in §\ref{sec:generation}).\footnote{We replaced USE \cite{cer2018universal} with SBERT MPNet embeddings which are perform better on the STS benchmark \cite{cer2017semeval} (75 $\rightarrow$ 88)}. We also use Recall@10 to compare the retrieved URLs for the ground-truth query and the generated query.\footnote{\url{https://serpapi.com/} provides an open API for a popular internet search engine.} We conduct experiments using an open-source T5-3B model \cite{raffel2020exploring} in its original form (referred to as `None'), by fine-tuning it on the natural QReCC training data and contrasting the results with those obtained from training on the auto-generated QReCC dataset. We use a batch size of 32, an Adam optimizer, a learning rate of 0.0001, and fine-tune it for 10,000 steps.

\paragraph{Results.} Results are presented in Table~\ref{tab:results_replaing}. We observe that by replacing human annotated data with auto generated data we were able to reach 90\%--95\% of the results with a set of the same size using the same model, demonstrating the efficacy of our \emph{q2d} approach in minimizing annotation labor and producing synthetic training data that is nearly as effective as human-annotated data.

\section{Extending Query Generation: Multi-Hop QA}
\label{sec:extending}
\begin{table}[!h]
\resizebox{\columnwidth}{!}{
\begin{tabular}{@{}lllllll@{}}
\toprule
                 & \multicolumn{2}{l}{\begin{tabular}[c]{@{}l@{}}SBERT \\ Similarity\end{tabular}} & \multicolumn{2}{l}{\begin{tabular}[c]{@{}l@{}}Rouge-1\\ Recall\end{tabular}} & \multicolumn{2}{l}{\begin{tabular}[c]{@{}l@{}}Search Results\\ Recall@10\end{tabular}} \\ \midrule
Model / Test Set & M                                      & B                                      & M                                     & B                                    & M                                          & B                                         \\ \midrule
WizInt           & 66                                     & 67                                     & 40                                    & 36                                   & 21                                         & 21                                        \\
BlenderBot3      & 62                                     & 69                                     & 32                                    & 35                                   & 19                                         & 24                                        \\
T5 (QReCC)       & 74                                     & 77                                     & 70                                    & 65                                   & 34                                         & 37                                        \\
PaLM 540B        & 88                                     & 82                                     & 81                                    & 69                                   & 52                                         & 41                                        \\
Flan-U-PaLM 540B & 89                                     & 82                                     & 83                                    & 68                                   & 57                                         & 39                                        \\ \midrule
\textbf{T5 (MuSiQue)}     & 97                                     & 91                                     & 94                                    & 80                                   & 75                                         & 54                                        \\ \bottomrule
\end{tabular}
}
\caption{Performance of language and dialogue models on query generation test sets is shown. `M'\`B' indicates results on MuSiQue auto-generated \ Bamboogle manually constructed dialogues. (QReCC) and (MuSiQue) indicate fine-tuning on a ``q2d'' dataset. Best results were achieved by models fine-tuned on MuSiQue auto-generated dialogue, which improved T5 results by 14\%-59\% on the human-annotated test.}
\label{tab:results_extending}
\end{table}


\begin{wrapfigure}{R}{0.5\columnwidth}
    \includegraphics[width=0.5\columnwidth]{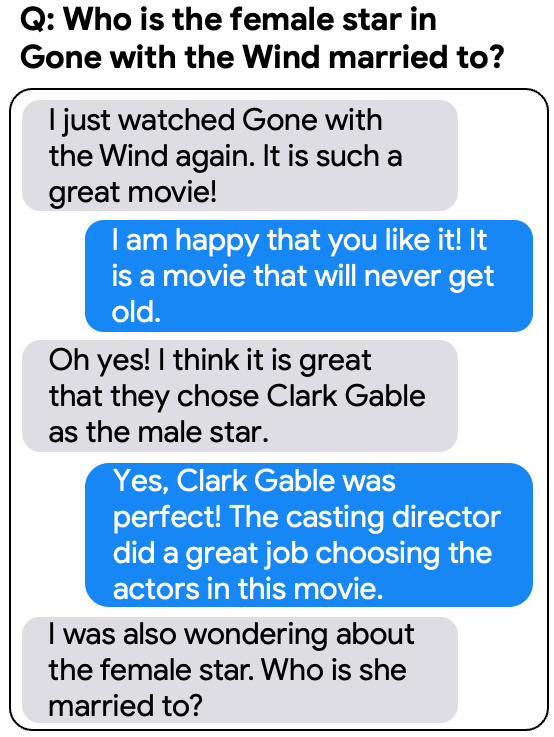}
  \caption{An example of auto-generated dialog, where we take a multi-hop question from MuSiQue and use \emph{q2d} to generate dialog in a conversational style with the same intent.}
\label{fig:example_multi_hop}
\end{wrapfigure}

This section shows that our method is effective as a benchmark and training signal that generalizes to human-annotated data. It is also flexible and able to adapt and improve for specific styles of dialog, even without annotated data. It allows us to create dialogs similar to a target domain and provide a fully labeled query-generation dataset. The generated data is useful for training and evaluation, as well as exploring model performance in new scenarios. We demonstrate this using a multi-hop question answering example.

\paragraph{Manual Dialog Construction.} We define a challenging test set for multi-hop dialogs by annotating the Bamboogle dataset \cite{press2022measuring}, which consists of 125 multi-hop human-constructed questions. We create dialogs that ask the same questions, with the user as the information seeker and the assistant as the information provider. The assistant should help the user obtain the information they are seeking, clarify any questions, and move the conversation forward without trying to mimic human-to-human interaction. An example from the generated dataset is presented in Figure~\ref{fig:example_multi_hop}. Full instructions, examples and annotated data can be found in the Appendix~\ref{sec:appendix}, Section~\ref{sec:bamboogle}, including examples with model predictions.

\paragraph{Dataset Generation.} We use our \emph{q2d} method as described in Section~\ref{sec:generation} to generate dialogs that ask multi-hop questions, using the MuSiQue dataset \cite{trivedi2021musique} as the base for the questions ($Q$). MuSiQue is a challenging multi-hop QA dataset that is partially auto-generated, so we generate dialogs from partially generated questions. This illustrates how we can use automatically generated data to improve on human-annotated data. We use seven few-shot examples ($S_q$ and $S_d$). As a result, we generate 3K train samples and 480 test samples. Full prompts, instructions and examples are available in Appendix~\ref{sec:appendix}, Section~\ref{sec:examples}.

\paragraph{Metrics.} The metrics used in this work are the same as those described in the previous section: Rouge-1 Recall, SBERT embedding semantic similarity, and Recall@10.

\paragraph{Models.} We evaluate several state-of-the-art language and dialog models. These include PaLM 540B \cite{chowdhery2022palm}, Flan-U-PaLM 540B \cite{chung2022scaling}, T5-3B \cite{raffel2020exploring}, BlenderBot3-3B \cite{shuster2022blenderbot}, WizInt Search Engine FiD \cite{lewis2019bart}
These models are used in a zero-shot setting, except for T5,
which is fine-tuned on the auto-generated MuSiQue dialogs in the same method presented in Section~\ref{sec:replacing}.  BlenderBot3 and WizInt are publicly available in Parlai \cite{miller2017parlai}, exact details and versions are described in Appendix~\ref{sec:appendix}, Section~\ref{sec:parlai_models}. More details on the instructions for zero-shot models can be found in the Appendix~\ref{sec:appendix}, Section~\ref{sec:appendix_prompts}.

\paragraph{Results.} Query generation results are presented in Table~\ref{tab:results_extending}.\footnote{We show relatively low scores with WizInt and BlenderBot3 that seem to be oriented in finding the topic query rather than concrete questions.} Qualitative examples with 
T5 model predictions are available in Appendix~\ref{sec:appendix}, Section~\ref{sec:examples}. 
The T5 model improves performance on the human-curated Bamboogle test by 14\%-59\% after fine-tuning on the auto-generated MuSiQue multi-hop dialogues. We show examples for it in Appendix~\ref{sec:appendix}, Section~\ref{sec:additional_examples}. This improvement also correlates with improvements on the auto-generated test set, indicating the effectiveness of our method for creating evaluation data. To conclude, our results show that our datasets are effective as a benchmark for query generation, as well as training data that generalizes to both auto-generated and human-annotated test sets.

\paragraph{Producing a Partially Decomposed Query.} Given a multi-hop dialog, query generation models may resolve partial information. For example, if a dialog asks ``How old is the current US president?'', a query generation model may produce ``How old is Joe Biden?'', which is correct at the time but may become outdated in the future, or may produce hallucinations. To prevent this, we can make two query generation calls (first to discover the current US president and then their age), decouple knowledge from executing \cite{borgeaud2021improving,parisi2022talm}, periodically update the model's weights, or disallow the model from making partial resolves. This will help ensure that the generated query remains accurate and relevant over time. The fine-tuning technique described in this section uses the last approach to avoid making assumptions about the current president’s age or
identity. 
\yoad{the goal of this paragraph is to justify why we choose fine-tunning approach?}\yonatan{Let's talk about it. This is a `defense' paragraph. Potentially we can eliminate it, but this paragraph shows that the current metrics are optimized to reward the complex query on account of a simple query that might be also correct. That's also some kind of limitation (or feature) of our approach}


\section{Intrinsic Evaluation: Naturalness, Factuality and Correctness}
\label{sec:analysis}
In this section we perform a thorough analysis of the generated dialogs, focusing on the QReCC NQ dataset which contains human annotated dialogs, and evaluate their naturalness (§\ref{sec:naturalness}), factuality (§\ref{sec:factuality}) and correctness (§\ref{sec:correctness}). 

\subsection{Naturalness: Humans Struggle to Distinguish Synthetic Dialogs from Natural}
\label{sec:naturalness}
We define a human-evaluation task to distinguish between naturally generated dialogs and auto-generated dialogs. We sample 100 annotated dialogs from QReCC NQ \cite{anantha2020open} and mix them with 100 dialogs we generated. The annotators, who are not authors of the paper and have a STEM degree,  were asked to mark 1 if the dialog seems to be generated by a machine, and 0 otherwise.\footnote{Full instructions to the annotators are provided in Appendix~\ref{sec:appendix}, Section~\ref{sec:human_evaluation_instructions}} The labels were hidden. We use three annotators for each sample and select their majority vote as the final answer. The results show that the majority vote achieved a success rate of 50.5\%, while the random chance is 50\%. All individual annotators achieved between 50\%--55\% in this task. In 26\% of the cases there is a full agreement between all three annotators. When all agreed, the result improves to 51.9\%, which is still close to random chance. These results indicate that humans struggle to differentiate between natural and auto-generated dialogs. This suggests that the auto-generated dialogs are of high quality and are similar to human annotations, and can be used in place of human-generated dialogs in certain situations, saving time and resources. 

\subsection{Factuality: Generated and Human-Annotated Answers Perform Similarly in Query Generation}
\label{sec:factuality}
\begin{figure}[!tb]
    \includegraphics[width=\columnwidth]{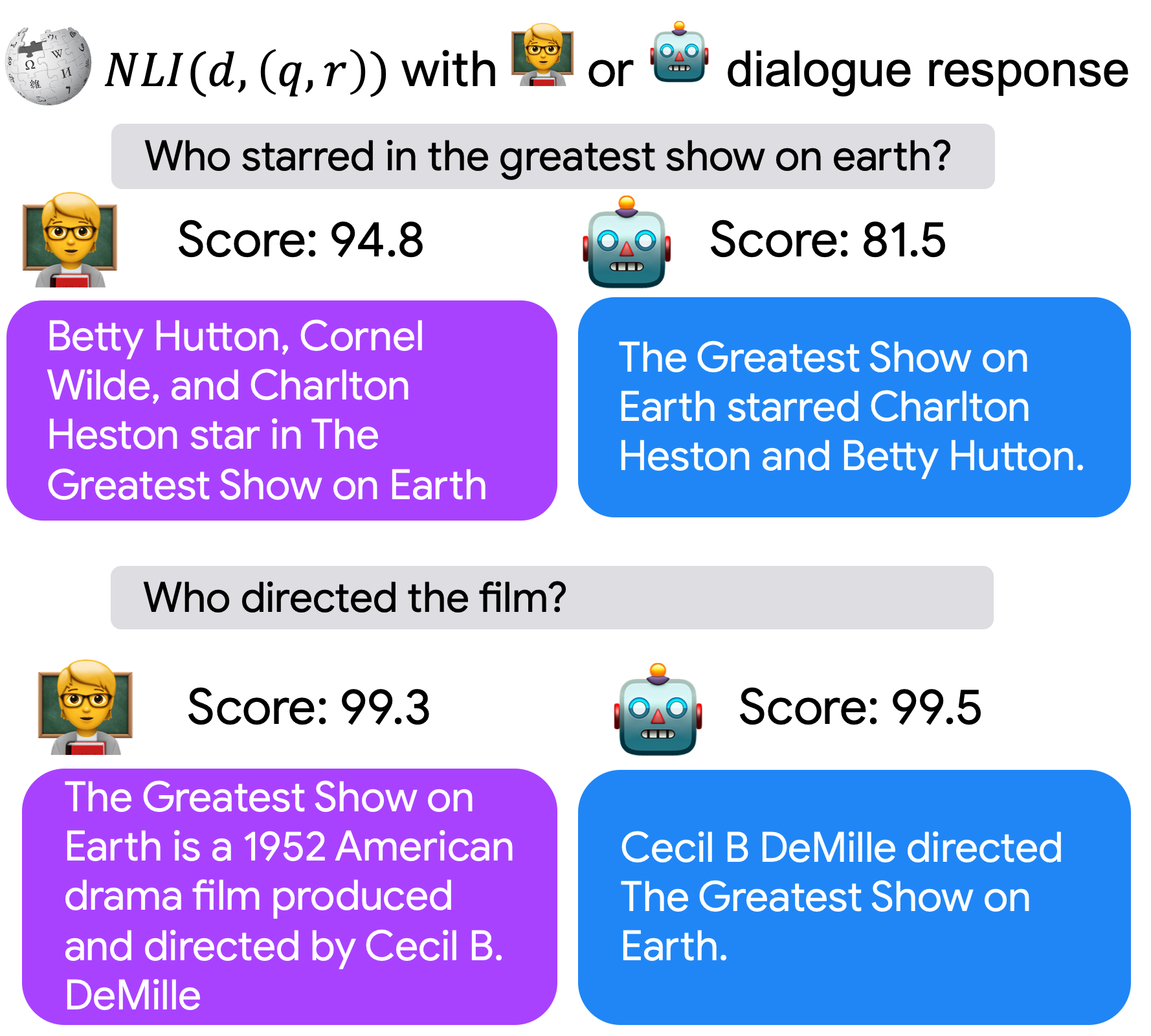}\\
    \caption{Illustration of the response factuality evaluation. For each turn, we produce a response with PaLM, and compare the generated response to the human annotated response. We use an NLI model to score whether the response answers the question (``Hypothesis: The answer to the question \{q\} is \{r\}'') according to the Wikipedia document $d$ used by the human annotator in the ground-truth response generation (``Premise: \{d\}''). In the first response there is a lower score for the PaLM response because it misses the mention of `Cornel Wilde' that appears in the \href{https://en.wikipedia.org/wiki/The_Greatest_Show_on_Earth_(film)}{document summary}.
    }
    \label{fig:answers_factuality}
\end{figure}

The q2d method generates a dialog by starting with a query and generating a series of related questions and answers. However, since the intermediate answers are generated by a large language model, there is a chance that they may be factually correct or incorrect. This raises the following questions. 
(1) Are the intermediate answers factually correct?
(2) How does the factuality of the generated answers affect the results of downstream tasks?

We replace all human annotated answers in the QReCC NQ training split with PaLM generated answers. To produce PaLM answers, we use a few-shot prompt, where the input is the original dialog ending in a question, and the output is the PaLM response. An example is provided in Figure~\ref{fig:answers_factuality}.

\paragraph{Intermediate Answers Factuality According to Automatic Metrics and Human Raters.} To answer the first question, we evaluate the factual correctness of the generated answers by using an NLI \cite{dagan2005pascal} model presented by \citet{honovich2021q}. We take the question (``q''), the response (``r'') that may be the ground-truth annotated response or the generated response, and the Wikipedia document (``d'') summary available in QReCC dataset. We construct the following NLI instance: ``premise: \{d\} hypothesis: The answer to the question \{q\} is \{r\}' and produce NLI scores for the ground-truth responses vs. the generated responses. Figure~\ref{fig:answers_factuality} illustrates our process. The average NLI scores for the human responses are 62\%, and for the PaLM responses is 38\%. However, this measure is biased towards the human responses since we measure it with the Wikipedia document that was used to generate the answer. PaLM might also produce a correct answer, that is just not written in the same exact words in Wikipedia. To test this, we conducted an annotation task with an annotator that is not a part of the paper authors. The annotator was presented with a 50 samples of dialog, query, and two options: A and B. One of the options was the original answer and the other was the generated answer. The annotator's task was to mark 0/1 for each answer indicating whether it was factual and relevant for the question. The results are that PaLM responses were marked as correct in 82\% of the cases, compared to 93\% correctness of the human responses. This result indicates the factuality and relevancy of the generated responses. 

\paragraph{For Query Generation, Generated Answers Perform Similar to Human-Annotated.} To answer the second question, we replace all of the human annotated answers with automatically generated answers, receiving a semi-auto-generated training set with the same structure and same annotated questions, but with PaLM generated dialogs. Then we train a T5-3B \cite{raffel2020exploring} model on the human annotated and the semi-auto-generated version and compare the results. For example in Figure~\ref{fig:answers_factuality}, the semi-auto-generated dialog is the one with the answers on the right side. We train the same way as we presented in Section~\ref{sec:replacing}. The results are 86.6\% Rouge-1 Recall with the semi auto-generated training set, only a small drop (1.5\%) from the results of the model trained on the natural data, indicating that although PaLM sometimes (<48\%) produce in-factual responses, it only has negligible effect on the query generation task.

\subsection{Correctness: Generated Datasets Labels are Accurate, and Strict Filtering Improves Results}
\label{sec:correctness}
\begin{figure}[!tb]
    \includegraphics[width=\columnwidth]{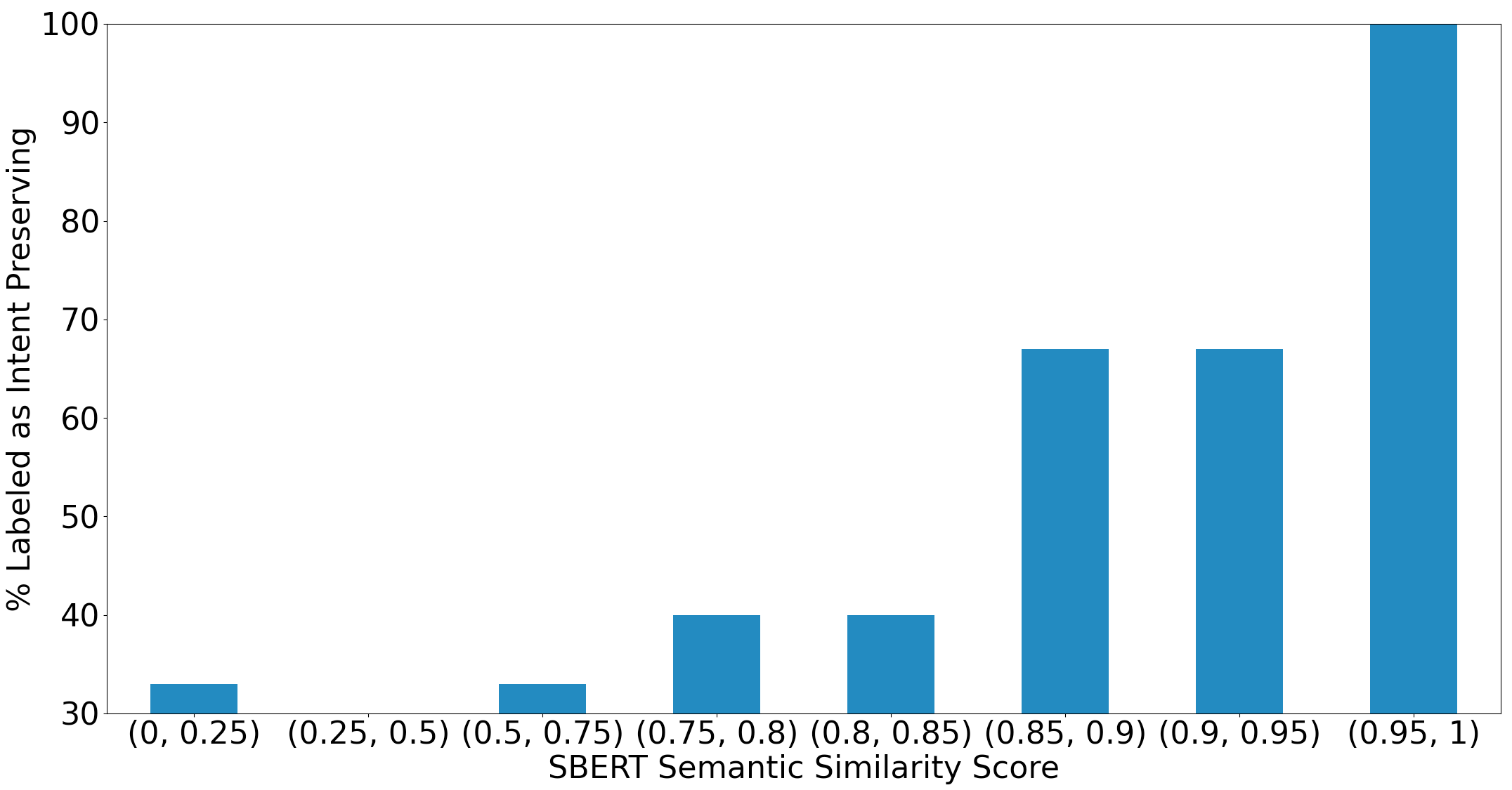}\\
    \caption{Intent-preserving annotation task results. The proportion of samples that were annotated as intent-preserving increases with the semantic similarity score.
    }
\label{fig:chart_intent_preserving}
\end{figure}

\begin{table}[!ht]
\resizebox{\columnwidth}{!}{
\begin{tabular}{@{}lll@{}}
\toprule
\begin{tabular}[c]{@{}l@{}}SBERT Similarity \\ Filtering Threshold\end{tabular} & \begin{tabular}[c]{@{}l@{}}Filtering \\ Proportion\end{tabular} & \begin{tabular}[c]{@{}l@{}}Query Generation\\ Rouge-1 Recall\end{tabular} \\ \midrule
0                                                                               & 0                       & 68                                                                           \\
0.25                                                                            & 6                       & 69                                                                           \\
0.5                                                                             & 16                      & 72                                                                           \\
0.75                                                                            & 37                      & 74                                                                           \\
0.8                                                                             & 44                      & 76                                                                           \\
0.9                                                                             & 62                      & 79                                                                           \\
0.95                                                                            & 72                      & 81                                                                           \\
0.99                                                                            & 84                      & 83                                                                           \\ 
\textbf{0.999}                                                                           & \textbf{88}                      & \textbf{84}                                                                           \\ \bottomrule
\end{tabular}}
\caption{Reversed queries similarity filter. The similarity is measured between the original query $q$ and the reversed query $q'$ predicted with the few-shot model $q' \gets M_{fs}(S_d, I_r, dialogue)$. The higher the filter threshold (strict filter), the better the results. }
\label{tab:filtering_ablations}
\end{table}

Our main filter measures a similarity between the original query and the reversed query $sim(q,q')$ and compare it to a threshold $T_{query}$. We measure its effect in human-evaluation and automatic ablation studies. Both experiments indicate the label correctness for the task of predicting the query from a dialog and the value of stricter filtering threshold.  

\paragraph{Humans Find that Dialogs Generated by Queries Have the Same Intent.}
We define a human annotation task to determine whether the dialogs are intent-preserving. Annotators were asked to mark 1 if the dialog is intent-preserving, and 0 otherwise.\footnote{Full instructions to the annotators are provided in Appendix~\ref{sec:appendix}, Section~\ref{sec:human_evaluation_instructions}} We use three annotators for each sample, and select their majority vote as the final answer. We follow the notation suggested by \citep{groenendijk1984studies} about entailment between questions: an interrogative $q$ entails another $d$ iff every proposition that answers $q$ answers $d$ as well \cite{jiang2022investigating}. Here, $q$ stands for a question and $d$ stands for an information-seeking dialog. We defined eight SBERT semantic similarity score buckets, with 15 in each, covering all similarities between 0 and 100. Results are presented in Figure~\ref{fig:chart_intent_preserving}. All three annotators agree in 88\% of the cases. The proportion of intent-preserving annotations grows according to the SBERT semantic similarity score, with a strong gain between 0.95 and 1, the only bucket with 100\% intent-preserving annotations. Accordingly, we only select samples that have generated queries very similar to the original query ($\ge$0.99) in the filtering step. 

\paragraph{Strict Filtering Leads to Higher Quality Data, Resulting in Improved Downstream Results.} We measure different thresholds tested on an evaluation set of 1,000 instances we generated from other train queries. We also add another filtering method based on an NLI \cite{dagan2005pascal} model, given a dialog ``d'' and a question ``q'', we construct the following NLI sample: ``premise: \{d\} hypothesis: The dialog asks the question \{q\}'', with different thresholds. 
Results are presented in Table~\ref{tab:filtering_ablations}. We report the Rouge-1 recall on the evaluation set. We see that performance increases as the reversed similarity threshold rises, and with a clear trade-off with the filtering proportion. The more data we generate, we are able to apply a more strict filtering, receiving higher quality data, that leads to better results.\footnote{The high filtering proportion means that we simply need to generate more data (requiring more compute time) in order to achieve a dataset of the same size without filtering.} We produced four options for the NLI-based method, with thresholds ranging from 0.65 to 0.82, and above it filtered too much data (below the goal of 13K). The max performance for the 0.82 threshold group is 70\%, much lower than the alternative reverse queries filter. 

\section{Related Work}
Our work relates to data generation, query generation for search-based models, and information retrieval datasets.

\paragraph{Data Generation} Several works have used large language models for data generation. \citet{dai2022promptagator} applies this technique to information retrieval, creating retrievers based on generated data that generate queries given the document. Their method involves round-consistency filtering using a large language model, a method similar to reverse translation. In the context of dialog generation, Dialog Inpaintint \cite{dai2022dialog} starts from a document and generates a dialog. Our approach focuses on generating dialogs from queries, which allows us to leverage the availability of existing QA datasets. This enables us to create information-seeking dialogs with the same intent as the original questions, along with automatically generated labels for the queries and answers. 

\paragraph{Search Based Query Generation}
dialog models like LaMDA and BlenderBot use search APIs to generate factual responses. Training and evaluation data for such models is obtained mostly with human annotated data. Previous works \cite{shuster2022blenderbot,thoppilan2022lamda,komeili2021internet} evaluated only the end-to-end dialog response without evaluating the generated query. The evaluation was primarily based on automated metrics of perplexity and F1, or with human annotations assessing whether the model response is sensible, specific, and interesting (SSI), or whether it is correct, engaging, and consistent. The evaluated dialogs were general, not necessarily information-seeking. The focus of this paper is on the query generation task for information-seeking dialogs, with a concrete question and an expected response.

\paragraph{Question Rewrite} Works like QReCC \cite{anantha2020open}, Question
Answering in Context (QuAC) \cite{choi2018quac}, TREC Conversational Assistant Track (CAsT) \cite{dalton2020cast}, QuAC and CANARD \cite{elgohary2019can} in the information retrieval domain use human annotated data, that mostly contain follow-up dialogs, questions followed by answers. Our work focuses on the application of dialog models like LaMDA and BlenderBot, which often involve the use of less formal language and more human-like conversations. The need for a variety of query generation datasets has motivated us to develop an automatic method for generating dialogs for the query generation task, with a range of different styles and skills required.

\section{Conclusions}

We introduced \emph{q2d}, a data generation pipeline that produces dialogs based on questions. We demonstrated that our method can replace human-annotated data to train query-generation models, and to create effective, natural, factual, and accurate evaluation and training data in new domains, even when no existing dialogue data is available.

\section{Acknowledgements}
\yonatan{Did we missed anyone?}
We would like to thank Gabriel Stanovsky, Roy Schwartz, Michael Elhadad, Alon Jacovi, Amir Globerson, Avinatan Hassidim, Douglas Eck, Itay Itzhak, Nitzan Bitton-Guetta, Netta Madvil, Michael Guzman and Ofir Press for their valuable feedback. We would also like to thank Uli Reuckert, and the Luna team, specifically Tamar Yakar, Avia Aharon, Oded Elyada, Stav Ginzburg and Yael Karov.  

\bibliography{anthology,custom}

\begin{thebibliography}{36}
\expandafter\ifx\csname natexlab\endcsname\relax\def\natexlab#1{#1}\fi

\bibitem[{Anantha et~al.(2020)Anantha, Vakulenko, Tu, Longpre, Pulman, and
  Chappidi}]{anantha2020open}
Raviteja Anantha, Svitlana Vakulenko, Zhucheng Tu, Shayne Longpre, Stephen
  Pulman, and Srinivas Chappidi. 2020.
\newblock Open-domain question answering goes conversational via question
  rewriting.
\newblock \emph{arXiv preprint arXiv:2010.04898}.

\bibitem[{Borgeaud et~al.(2021)Borgeaud, Mensch, Hoffmann, Cai, Rutherford,
  Millican, Driessche, Lespiau, Damoc, Clark et~al.}]{borgeaud2021improving}
Sebastian Borgeaud, Arthur Mensch, Jordan Hoffmann, Trevor Cai, Eliza
  Rutherford, Katie Millican, George van~den Driessche, Jean-Baptiste Lespiau,
  Bogdan Damoc, Aidan Clark, et~al. 2021.
\newblock Improving language models by retrieving from trillions of tokens.
\newblock \emph{arXiv preprint arXiv:2112.04426}.

\bibitem[{Brown et~al.(2020)Brown, Mann, Ryder, Subbiah, Kaplan, Dhariwal,
  Neelakantan, Shyam, Sastry, Askell et~al.}]{brown2020language}
Tom Brown, Benjamin Mann, Nick Ryder, Melanie Subbiah, Jared~D Kaplan, Prafulla
  Dhariwal, Arvind Neelakantan, Pranav Shyam, Girish Sastry, Amanda Askell,
  et~al. 2020.
\newblock Language models are few-shot learners.
\newblock \emph{Advances in neural information processing systems},
  33:1877--1901.

\bibitem[{Cer et~al.(2017)Cer, Diab, Agirre, Lopez-Gazpio, and
  Specia}]{cer2017semeval}
Daniel Cer, Mona Diab, Eneko Agirre, Inigo Lopez-Gazpio, and Lucia Specia.
  2017.
\newblock Semeval-2017 task 1: Semantic textual similarity-multilingual and
  cross-lingual focused evaluation.
\newblock \emph{arXiv preprint arXiv:1708.00055}.

\bibitem[{Cer et~al.(2018)Cer, Yang, Kong, Hua, Limtiaco, John, Constant,
  Guajardo-Cespedes, Yuan, Tar et~al.}]{cer2018universal}
Daniel Cer, Yinfei Yang, Sheng-yi Kong, Nan Hua, Nicole Limtiaco, Rhomni~St
  John, Noah Constant, Mario Guajardo-Cespedes, Steve Yuan, Chris Tar, et~al.
  2018.
\newblock Universal sentence encoder for english.
\newblock In \emph{Proceedings of the 2018 conference on empirical methods in
  natural language processing: system demonstrations}, pages 169--174.

\bibitem[{Choi et~al.(2018)Choi, He, Iyyer, Yatskar, Yih, Choi, Liang, and
  Zettlemoyer}]{choi2018quac}
Eunsol Choi, He~He, Mohit Iyyer, Mark Yatskar, Wen-tau Yih, Yejin Choi, Percy
  Liang, and Luke Zettlemoyer. 2018.
\newblock Quac: Question answering in context.
\newblock \emph{arXiv preprint arXiv:1808.07036}.

\bibitem[{Chowdhery et~al.(2022)Chowdhery, Narang, Devlin, Bosma, Mishra,
  Roberts, Barham, Chung, Sutton, Gehrmann et~al.}]{chowdhery2022palm}
Aakanksha Chowdhery, Sharan Narang, Jacob Devlin, Maarten Bosma, Gaurav Mishra,
  Adam Roberts, Paul Barham, Hyung~Won Chung, Charles Sutton, Sebastian
  Gehrmann, et~al. 2022.
\newblock Palm: Scaling language modeling with pathways.
\newblock \emph{arXiv preprint arXiv:2204.02311}.

\bibitem[{Chung et~al.(2022)Chung, Hou, Longpre, Zoph, Tay, Fedus, Li, Wang,
  Dehghani, Brahma et~al.}]{chung2022scaling}
Hyung~Won Chung, Le~Hou, Shayne Longpre, Barret Zoph, Yi~Tay, William Fedus,
  Eric Li, Xuezhi Wang, Mostafa Dehghani, Siddhartha Brahma, et~al. 2022.
\newblock Scaling instruction-finetuned language models.
\newblock \emph{arXiv preprint arXiv:2210.11416}.

\bibitem[{Dagan et~al.(2005)Dagan, Glickman, and Magnini}]{dagan2005pascal}
Ido Dagan, Oren Glickman, and Bernardo Magnini. 2005.
\newblock The pascal recognising textual entailment challenge.
\newblock In \emph{Machine learning challenges workshop}, pages 177--190.
  Springer.

\bibitem[{Dai et~al.(2022{\natexlab{a}})Dai, Chaganty, Zhao, Amini, Rashid,
  Green, and Guu}]{dai2022dialog}
Zhuyun Dai, Arun~Tejasvi Chaganty, Vincent~Y Zhao, Aida Amini, Qazi~Mamunur
  Rashid, Mike Green, and Kelvin Guu. 2022{\natexlab{a}}.
\newblock Dialog inpainting: Turning documents into dialogs.
\newblock In \emph{International Conference on Machine Learning}, pages
  4558--4586. PMLR.

\bibitem[{Dai et~al.(2022{\natexlab{b}})Dai, Zhao, Ma, Luan, Ni, Lu, Bakalov,
  Guu, Hall, and Chang}]{dai2022promptagator}
Zhuyun Dai, Vincent~Y Zhao, Ji~Ma, Yi~Luan, Jianmo Ni, Jing Lu, Anton Bakalov,
  Kelvin Guu, Keith~B Hall, and Ming-Wei Chang. 2022{\natexlab{b}}.
\newblock Promptagator: Few-shot dense retrieval from 8 examples.
\newblock \emph{arXiv preprint arXiv:2209.11755}.

\bibitem[{Dalton et~al.(2020)Dalton, Xiong, Kumar, and Callan}]{dalton2020cast}
Jeffrey Dalton, Chenyan Xiong, Vaibhav Kumar, and Jamie Callan. 2020.
\newblock Cast-19: A dataset for conversational information seeking.
\newblock In \emph{Proceedings of the 43rd International ACM SIGIR Conference
  on Research and Development in Information Retrieval}, pages 1985--1988.

\bibitem[{Dziri et~al.(2021)Dziri, Rashkin, Linzen, and
  Reitter}]{dziri2021evaluating}
Nouha Dziri, Hannah Rashkin, Tal Linzen, and David Reitter. 2021.
\newblock Evaluating groundedness in dialogue systems: The begin benchmark.
\newblock \emph{arXiv preprint arXiv:2105.00071}.

\bibitem[{Elgohary et~al.(2019)Elgohary, Peskov, and
  Boyd-Graber}]{elgohary2019can}
Ahmed Elgohary, Denis Peskov, and Jordan Boyd-Graber. 2019.
\newblock Can you unpack that? learning to rewrite questions-in-context.
\newblock \emph{Can You Unpack That? Learning to Rewrite Questions-in-Context}.

\bibitem[{Glaese et~al.(2022)Glaese, McAleese, Tr{\k{e}}bacz, Aslanides,
  Firoiu, Ewalds, Rauh, Weidinger, Chadwick, Thacker
  et~al.}]{glaese2022improving}
Amelia Glaese, Nat McAleese, Maja Tr{\k{e}}bacz, John Aslanides, Vlad Firoiu,
  Timo Ewalds, Maribeth Rauh, Laura Weidinger, Martin Chadwick, Phoebe Thacker,
  et~al. 2022.
\newblock Improving alignment of dialogue agents via targeted human judgements.
\newblock \emph{arXiv preprint arXiv:2209.14375}.

\bibitem[{Groenendijk and Stokhof(1984)}]{groenendijk1984studies}
Jeroen Antonius~Gerardus Groenendijk and Martin Johan~Bastiaan Stokhof. 1984.
\newblock \emph{Studies on the Semantics of Questions and the Pragmatics of
  Answers}.
\newblock Ph.D. thesis, Univ. Amsterdam.

\bibitem[{Gupta et~al.(2021)Gupta, Wu, Liu, and Xiong}]{gupta2021dialfact}
Prakhar Gupta, Chien-Sheng Wu, Wenhao Liu, and Caiming Xiong. 2021.
\newblock Dialfact: A benchmark for fact-checking in dialogue.
\newblock \emph{arXiv preprint arXiv:2110.08222}.

\bibitem[{Honovich et~al.(2022)Honovich, Aharoni, Herzig, Taitelbaum,
  Kukliansy, Cohen, Scialom, Szpektor, Hassidim, and Matias}]{honovich2022true}
Or~Honovich, Roee Aharoni, Jonathan Herzig, Hagai Taitelbaum, Doron Kukliansy,
  Vered Cohen, Thomas Scialom, Idan Szpektor, Avinatan Hassidim, and Yossi
  Matias. 2022.
\newblock True: Re-evaluating factual consistency evaluation.
\newblock \emph{arXiv preprint arXiv:2204.04991}.

\bibitem[{Honovich et~al.(2021)Honovich, Choshen, Aharoni, Neeman, Szpektor,
  and Abend}]{honovich2021q}
Or~Honovich, Leshem Choshen, Roee Aharoni, Ella Neeman, Idan Szpektor, and Omri
  Abend. 2021.
\newblock $ q^{\textnormal{2}} $: Evaluating factual consistency in
  knowledge-grounded dialogues via question generation and question answering.
\newblock \emph{arXiv preprint arXiv:2104.08202}.

\bibitem[{Jiang and de~Marneffe(2022)}]{jiang2022investigating}
Nan-Jiang Jiang and Marie-Catherine de~Marneffe. 2022.
\newblock Investigating reasons for disagreement in natural language inference.
\newblock \emph{arXiv preprint arXiv:2209.03392}.

\bibitem[{Komeili et~al.(2021)Komeili, Shuster, and
  Weston}]{komeili2021internet}
Mojtaba Komeili, Kurt Shuster, and Jason Weston. 2021.
\newblock Internet-augmented dialogue generation.
\newblock \emph{arXiv preprint arXiv:2107.07566}.

\bibitem[{Kwiatkowski et~al.(2019)Kwiatkowski, Palomaki, Redfield, Collins,
  Parikh, Alberti, Epstein, Polosukhin, Devlin, Lee
  et~al.}]{kwiatkowski2019natural}
Tom Kwiatkowski, Jennimaria Palomaki, Olivia Redfield, Michael Collins, Ankur
  Parikh, Chris Alberti, Danielle Epstein, Illia Polosukhin, Jacob Devlin,
  Kenton Lee, et~al. 2019.
\newblock Natural questions: a benchmark for question answering research.
\newblock \emph{Transactions of the Association for Computational Linguistics},
  7:453--466.

\bibitem[{Lee et~al.(2018)Lee, Firat, Agarwal, Fannjiang, and
  Sussillo}]{lee2018hallucinations}
Katherine Lee, Orhan Firat, Ashish Agarwal, Clara Fannjiang, and David
  Sussillo. 2018.
\newblock Hallucinations in neural machine translation.

\bibitem[{Lewis et~al.(2019)Lewis, Liu, Goyal, Ghazvininejad, Mohamed, Levy,
  Stoyanov, and Zettlemoyer}]{lewis2019bart}
Mike Lewis, Yinhan Liu, Naman Goyal, Marjan Ghazvininejad, Abdelrahman Mohamed,
  Omer Levy, Ves Stoyanov, and Luke Zettlemoyer. 2019.
\newblock Bart: Denoising sequence-to-sequence pre-training for natural
  language generation, translation, and comprehension.
\newblock \emph{arXiv preprint arXiv:1910.13461}.

\bibitem[{Lin(2004)}]{lin2004rouge}
Chin-Yew Lin. 2004.
\newblock Rouge: A package for automatic evaluation of summaries.
\newblock In \emph{Text summarization branches out}, pages 74--81.

\bibitem[{Maynez et~al.(2020)Maynez, Narayan, Bohnet, and
  McDonald}]{maynez2020faithfulness}
Joshua Maynez, Shashi Narayan, Bernd Bohnet, and Ryan McDonald. 2020.
\newblock On faithfulness and factuality in abstractive summarization.
\newblock \emph{arXiv preprint arXiv:2005.00661}.

\bibitem[{Miller et~al.(2017)Miller, Feng, Fisch, Lu, Batra, Bordes, Parikh,
  and Weston}]{miller2017parlai}
Alexander~H Miller, Will Feng, Adam Fisch, Jiasen Lu, Dhruv Batra, Antoine
  Bordes, Devi Parikh, and Jason Weston. 2017.
\newblock Parlai: A dialog research software platform.
\newblock \emph{arXiv preprint arXiv:1705.06476}.

\bibitem[{Nie et~al.(2020)Nie, Williamson, Bansal, Kiela, and
  Weston}]{nie2020like}
Yixin Nie, Mary Williamson, Mohit Bansal, Douwe Kiela, and Jason Weston. 2020.
\newblock I like fish, especially dolphins: Addressing contradictions in
  dialogue modeling.
\newblock \emph{arXiv preprint arXiv:2012.13391}.

\bibitem[{Parisi et~al.(2022)Parisi, Zhao, and Fiedel}]{parisi2022talm}
Aaron Parisi, Yao Zhao, and Noah Fiedel. 2022.
\newblock Talm: Tool augmented language models.
\newblock \emph{arXiv preprint arXiv:2205.12255}.

\bibitem[{Press et~al.(2022)Press, Zhang, Min, Schmidt, Smith, and
  Lewis}]{press2022measuring}
Ofir Press, Muru Zhang, Sewon Min, Ludwig Schmidt, Noah~A Smith, and Mike
  Lewis. 2022.
\newblock Measuring and narrowing the compositionality gap in language models.
\newblock \emph{arXiv preprint arXiv:2210.03350}.

\bibitem[{Raffel et~al.(2020)Raffel, Shazeer, Roberts, Lee, Narang, Matena,
  Zhou, Li, Liu et~al.}]{raffel2020exploring}
Colin Raffel, Noam Shazeer, Adam Roberts, Katherine Lee, Sharan Narang, Michael
  Matena, Yanqi Zhou, Wei Li, Peter~J Liu, et~al. 2020.
\newblock Exploring the limits of transfer learning with a unified text-to-text
  transformer.
\newblock \emph{J. Mach. Learn. Res.}, 21(140):1--67.

\bibitem[{Shuster et~al.(2022{\natexlab{a}})Shuster, Komeili, Adolphs, Roller,
  Szlam, and Weston}]{shuster2022language}
Kurt Shuster, Mojtaba Komeili, Leonard Adolphs, Stephen Roller, Arthur Szlam,
  and Jason Weston. 2022{\natexlab{a}}.
\newblock Language models that seek for knowledge: Modular search \& generation
  for dialogue and prompt completion.
\newblock \emph{arXiv preprint arXiv:2203.13224}.

\bibitem[{Shuster et~al.(2022{\natexlab{b}})Shuster, Xu, Komeili, Ju, Smith,
  Roller, Ung, Chen, Arora, Lane et~al.}]{shuster2022blenderbot}
Kurt Shuster, Jing Xu, Mojtaba Komeili, Da~Ju, Eric~Michael Smith, Stephen
  Roller, Megan Ung, Moya Chen, Kushal Arora, Joshua Lane, et~al.
  2022{\natexlab{b}}.
\newblock Blenderbot 3: a deployed conversational agent that continually learns
  to responsibly engage.
\newblock \emph{arXiv preprint arXiv:2208.03188}.

\bibitem[{Thoppilan et~al.(2022)Thoppilan, De~Freitas, Hall, Shazeer,
  Kulshreshtha, Cheng, Jin, Bos, Baker, Du et~al.}]{thoppilan2022lamda}
Romal Thoppilan, Daniel De~Freitas, Jamie Hall, Noam Shazeer, Apoorv
  Kulshreshtha, Heng-Tze Cheng, Alicia Jin, Taylor Bos, Leslie Baker, Yu~Du,
  et~al. 2022.
\newblock Lamda: Language models for dialog applications.
\newblock \emph{arXiv preprint arXiv:2201.08239}.

\bibitem[{Trivedi et~al.(2021)Trivedi, Balasubramanian, Khot, and
  Sabharwal}]{trivedi2021musique}
Harsh Trivedi, Niranjan Balasubramanian, Tushar Khot, and Ashish Sabharwal.
  2021.
\newblock Musique: Multi-hop questions via single-hop question composition.
\newblock \emph{arXiv preprint arXiv:2108.00573}.

\bibitem[{Zhao et~al.(2020)Zhao, Cohen, and Webber}]{zhao2020reducing}
Zheng Zhao, Shay~B Cohen, and Bonnie Webber. 2020.
\newblock Reducing quantity hallucinations in abstractive summarization.
\newblock \emph{arXiv preprint arXiv:2009.13312}.

\end{thebibliography}
\bibliographystyle{acl_natbib}

\appendix

\section{Appendix}
\label{sec:appendix}

\subsection{Generated Examples}
\label{sec:examples}
\begin{figure*}[!ht]
\centering
\includegraphics[width=\textwidth]{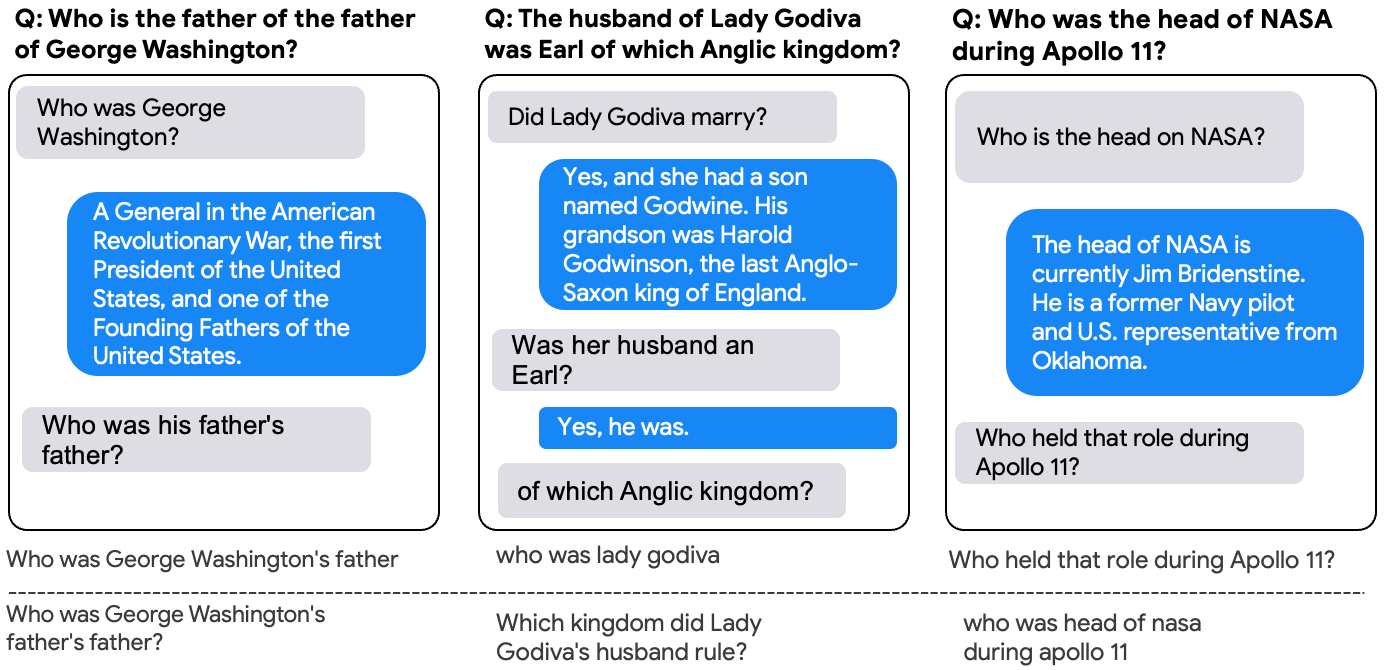}
\caption{Examples from the \textbf{human-annotated} dialogues for Bamboogle. The model predictions above/below the line are of T5, before/after fine-tuning on MuSiQue dialogues.}
\label{fig:examples_bamboogle}
\end{figure*}

\begin{figure*}[!ht]
\centering
\includegraphics[width=\textwidth]{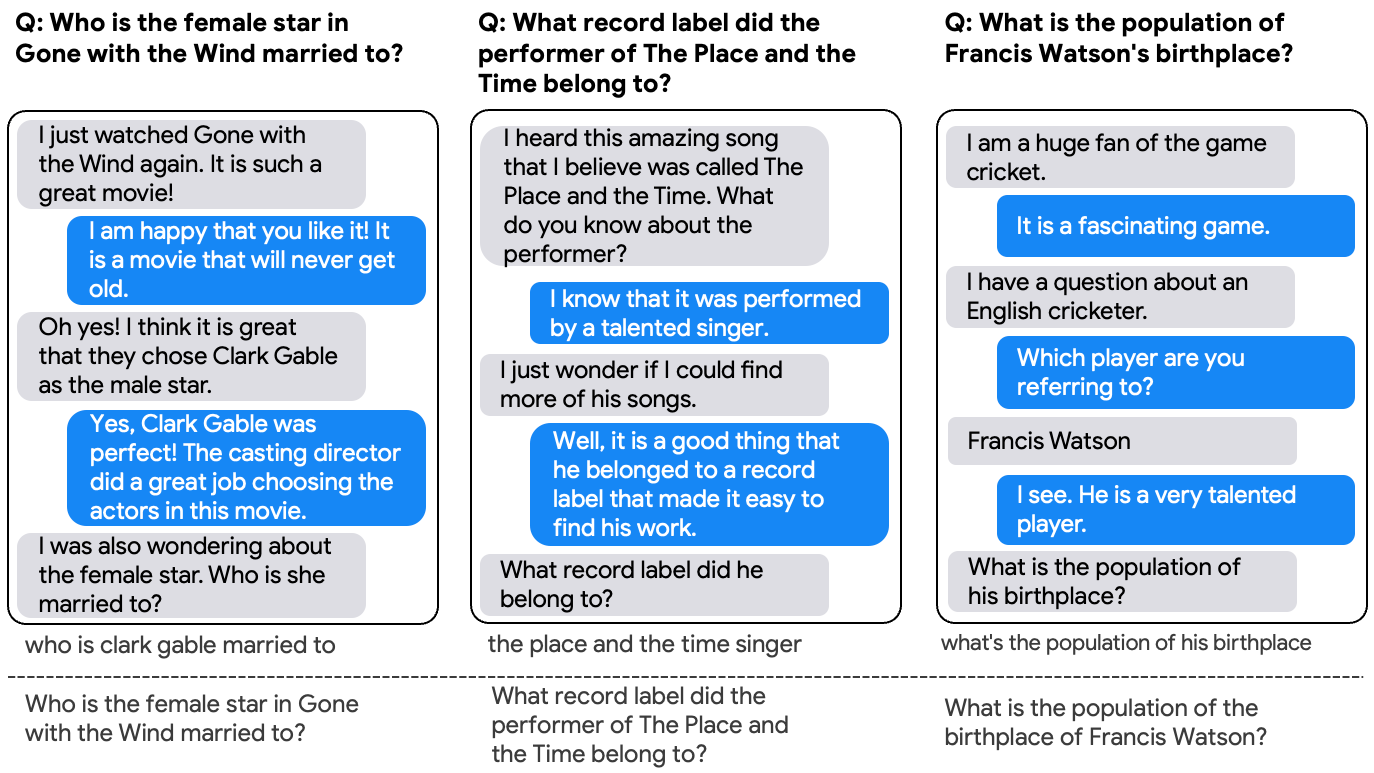}
\caption{Examples from the \textbf{auto-generated} dialogues for MuSiQue. The model predictions above/below the line are of T5, before/after fine-tuning on MuSiQue dialogues.}
\label{fig:examples_musique}
\end{figure*}

Figure~\ref{fig:examples_bamboogle} and Figure~\ref{fig:examples_musique} show \emph{cherry picked} examples of human-annotated / auto-generated examples from Bamboogle / MuSiQue. 

Tables~\ref{tab:random_examples_qrecc} and Tables~\ref{tab:random_examples_musique} show \emph{random} examples from the auto-generated QReCC / MuSiQue dialogs.

\begin{table}[!ht]
    \centering
    \small
    \caption{Random Examples from Auto-Generated QRECC}
    \label{tab:random_examples_qrecc}
    \begin{tabular}{|p{\linewidth}|} \toprule
    
Dialog: \\  
User: who plays the lead role in wish upon a star \\
Assistant: The cast of Wish Upon a Star include Katherine Heigl, Danielle Panabaker, and Mary Page Keller. \\
User: who plays haley \\ 
Question: who plays haley in wish upon a star? \\ 
Answer: Danielle Harris \\ 
\\
Dialog:\\
User: where is assam \\ 
Assistant: Assam is a state in India \\ 
User: what is the capital \\ 
Assistant: The capital of Assam is Dispur \\ 
User: who lead the first revolt against british in the state \\ 
Question: who lead the first revolt against british in assam?\\ 
Answer: Gomdhar Konwar \\ 
\\
Dialog:\\
User: when did last man standing air \\
Assistant: Last Man Standing premiered on ABC on October 11, 2011.  \\ 
User: what is the main character's name \\ 
Assistant: Tim Allen stars as Mike Baxter, the central character of the series. \\ 
User: what are the other character's names \\ 
Assistant: The other characters are Mike's wife and three daughters. \\ 
User: who plays cami harris \\
Question: who plays cami harris on last man standing?\\ 
Answer: Sarah Gilman \\ 
\\
Dialog:\\
User: who were the confederates in the civil war \\ 
Assistant: The Confederates in the American Civil War were the Confederate States of America, or the South. \\ 
User: who fought them \\
Question: who fought the confederates in the american civil war?\\ 
Answer: The Union \\ 
\\
Dialog:\\
User: when did adele webber first come on grey's anatomy
Assistant: The character of Adele Webber was first introduced in the episode ""Save Me"". \\ 
User: who plays her \\ 
Question: who plays adele webber on grey's anatomy?\\ 
Answer: Loretta Devine \\ 
\\
Dialog:\\
User: what did the effigy mounds look like \\ 
Assistant: The effigy mounds are earthen mounds built in the shapes of animals, humans, and other objects.  \\  
User: when were they discovered \\ 
Assistant: The first effigy mounds were discovered in the 1700s but were not identified as such until the late 1800s. \\ 
User: when were most of them built \\ 
Questions: when were most of the effigy mounds built? \\ 
Answer: 350 - 1300 CE \\ 
\\
\bottomrule
    \end{tabular}
\end{table}

\begin{table}[!ht]
    \centering
    \small
    \caption{Random Examples from Auto-Generated MuSiQue}
    \label{tab:random_examples_musique}
    \begin{tabular}{|p{\linewidth}|} \toprule
    
Dialog: \\  
User: I am looking for a movie to watch tonight. \\ 
Assistant: I think I can help you find a good movie. What are your preferences? \\ 
User: I like watching action movies. \\ 
Assistant: What about The Bourne Deception? It is a very popular action movie. \\
User: Who plays the main character of this movie? \\ 
Question: Who plays the main character of The Bourne Deception? \\ 
Answer: Matt Damon \\ 
\\
Dialog:\\
User: I just watched Charlie and the Chocolate Factory and I can’t get it out of my mind!\\
Assistant: What an amazing movie! I love it!\\
User: I have an argument with my sister about the spoilt character.\\
Assistant: That's the thing - there are so many characters in this movie that are not just black and white. They are complex characters. \\
User: I know the spoilt character is a boy but can you tell me who played him? \\ 
Question: Who played the spoilt character in Charlie and the Chocolate Factory?\\ 
Answer: Julie Dawn Cole \\ 
\\
Dialog:\\
User: I was watching the movie 'Dolores Claiborne' yesterday. It made me curious about the producer of the movie.\\
Assistant: It is the movie with Kathy Bates, right? \\ 
User: Yes, and I was wondering about the producer, do you know who his spouse is? \\ 
Question: Who is the spouse of the producer of Dolores Claiborne?\\ 
Answer: Helen Mirren \\ 
\\
Dialog:\\
User: I have this friend that is totally into football. \\ 
Assistant: I see. \\ 
User: He is a big fan of the Al Janoub Stadium. \\ 
Assistant: I am familiar with the Al Janoub Stadium. It is a very modern stadium, and its design is very unique. \\ 
User: I'm not surprised. I'm wondering what league includes the operating group of the stadium. \\ 
Question: What league includes the operating group of Al Janoub Stadium?\\
Answer: Qatar Stars League \\ 

\\
Dialog:\\
User: I am really into books these days. I just finished reading 'The Final Testament of the Holy Bible' and can't get it out of my mind.\\
Assistant: I can see how this book would be interesting. A lot of people think about books like that.\\ 
User: It is a really good book. I think that this book was published by a great publisher.\\ 
Assistant: Yes, it is. It is one of the best in the UK. \\ 
User: Do you know who founded the publisher of this book? \\ 
Questions: Who founded the publisher of The Final Testament of the Holy Bible? \\ 
Answer: Larry Gagosian \\ 
\\
\bottomrule
    \end{tabular}
\end{table}

\subsection{Filtering Examples}
\label{sec:filtering_examples}
\begin{table}[!ht]
    \centering
    \small
    \caption{Examples from the generated QReCC data with different SBERT similarity scores between original Query ($Q$) and the reversed Query ($R_Q$). The higher the similarity, the more the dialog's intent is the same as the original query. We took only dialogs with SBERT similarity $\ge0.999$.}
    \label{tab:examples_filtering}
    \begin{tabular}{|p{\linewidth}|} \toprule
    
SBERT Similarity: 0.999\\
Dialog:\\
User: who is the chairman of the joint chiefs of staff\\
Assistant: General Joseph Dunford is the current Chairman of the Joint Chiefs of Staff.\\
User: who does he advise\\
$Q$: who does the chairman of the joint chiefs of staff advise\\
$R_Q$: Who does the chairman of the joint chiefs of staff advise?\\
\\
SBERT Similarity: 0.75\\
Dialog:\\
User: who designed magic the gathering\\
Assistant: Richard Garfield is the creator of the Magic: The Gathering collectible card game.\\
User: who originally published the game\\
Assistant: Wizards of the Coast, a subsidiary of Hasbro, Inc. is the original publisher of Magic: The Gathering\\
User: who is the current publisher\\
$Q$: who created magic the gathering?\\
$R_Q$: who is the current publisher of the game Magic: The Gathering?\\
\\
SBERT Similarity: 0.5\\
Dialog:\\
User: what year did the Miami Hurricanes win the National Championship\\
Assistant: The Miami Hurricanes won the national championship in 2001.\\
User: was the team undefeated\\
Assistant: The Hurricanes finished the 2001 season 12–0.\\
User: who did they win against in the national championship game\\
$Q$: who did the Miaimi Hurricanes win against in 1983\\
$R_Q$: Who did the Miami Hurricanes play in the national championship game\\
\\
SBERT Similarity: 0.25\\
Dialog:\\
User: who won the golden glove award at the 2018 fifa world cup\\
Assistant: Thibaut Courtois was named the winner of the Golden Glove at the 2018 FIFA World Cup.\\
User: what team did he play for?\\
$Q$: who won the golden glove at the 2018 fifa world cup?\\
$R_Q$: what team did thibaut courtois play for?\\
\\
SBERT Similarity: 0\\
Dialog:\\
User: who inspired the book wonder by RJ Palacio\\
Assistant: The book Wonder by RJ Palacio was inspired by a childhood story of a boy with a severe facial difference.\\
User: what was the name of the boy\\
Assistant: The name of the boy was Julian.\\
User: what was his facial difference\\
Assistant: The boy had a craniofacial condition called Treacher Collins syndrome.\\
User: what is the condition\\
$Q$: what inspired the book Wonder\\
$R_Q$: what is Treacher Collins syndrome\\
\bottomrule
    \end{tabular}
\end{table}

Table~\ref{tab:examples_filtering} show examples of generated samples with different SBERT similarity scores used for filtering.

\subsection{Prompts and Instructions}
\label{sec:appendix_prompts}
Prompts (instruction + few-shot examples) for the auto-generated dialogs are presented in Table~\ref{tab:prompts_musique} for MuSiQue, and Table~\ref{tab:prompts_qrecc} for QReCC NQ. We show three prompts for each dataset due page page length, full prompts files are here: \href{https://drive.google.com/file/d/1Q2xxq0OeVdNdy28Zj4xRVLuRFk05M9Ua/view?usp=share_link}{MuSiQue}, \href{https://drive.google.com/file/d/1rM9dmvEGSbD7yCjtH9YRJj0E3fmPOaqG/view?usp=share_link}{QReCC}.

The reverse generation model uses the same examples but in reversed order (to generate a query given a dialog). 

The instruction for the reverse generation model and the PaLM zero-shot is: ``Given a dialog that asks an indirect question, extract the concrete question''.

The instructions for the Flan-U-PaLM-540B and are a bit more detailed (we found out it worked better for this instruction fine-tuned model): ``Your task is to reformulate the last conversational query in a dialog to a fully specified, context-independent query that can be effectively handled by Google Search.''.
\begin{table}[!ht]
    \centering
    \small
    \caption{MuSiQue Auto-Generated Dialogues Prompts}
    \label{tab:prompts_musique}

    \begin{tabular}{|p{\linewidth}|} \toprule

Write a dialog between an automated assistant and a user, and the dialog should indirectly ask the initial question you received. \\ 
\\
\\
Question: When was the institute that owned The Collegian founded? \\ 
Dialog: \\  
User: I have this homework that I need to submit in my history of the journalism course. Can you help me find out some of the details? \\ 
Assistant: Sure, I am here to help \\ 
User: I am working together with my friend Darren. We are looking into different newspapers, focusing on the powers that own them. I'm currently looking into The Collegian \\ 
Assistant: I can find out about The Collegian. It is actually owned by an educational institute. \\ 
User: When was the institute founded? \\ 
\\
\\
Question: What city is the person who broadened the doctrine of philosophy of language from?\\ 
Dialog:\\ 
User: I am conducting some research in the area of doctrine of philosophy of language\\ 
Assistant: I see. It is a fascinating sub-field of linguistics. It developed in quite an interesting process.\\ 
User: I know that it was broadened by some important philosopher\\ 
Assistant: Indeed, a famous german philosopher broadened it by focusing on language and its limits and ambiguities
\\ User: What city was he from?

\\
\\
Question: Who employs the person who wrote the book Animal Liberation?\\ 
Dialog:\\
User: I just finished reading the book Animal Liberation. To be honest, I can’t get the book out of my mind!
Assistant: Indeed, some people find animal liberation extremely important. Others care a lot more about humans and do not view this topic as very important.\\\
User: I wish I could write such influential books one day. I am not sure if my company would even let me do that.\\
Assistant: Well, fortunately for the Australian writer of the book, he works for a company that allowed him to fulfill himself and publish the book\\
User: Which company was that \\ \bottomrule

    \end{tabular}
\end{table}

\begin{table}[!ht]
    \centering
    \small
    \caption{QRECC NQ Auto-Generated Dialogues Prompts}
    \label{tab:prompts_qrecc}

    \begin{tabular}{|p{\linewidth}|} \toprule

Write a dialog between an automated assistant and a user, and the dialog should indirectly ask the initial question you received. \\ 
\\
\\
Question: Why was the great wall built?\\
Dialog:\\
User: where is the the great wall of china located\\ 
Assistant: The Great Wall of China is built across the historical northern borders of China.\\
User: how long is the wall\\
Assistant: The Great Wall is 21,196 km (13,171 mi).\\
User: why was the wall built\\
\\
\\
Question: Which U.S. states produce the most crued oil?\\
Dialog:\\
User: What kind of oil is in North Dakota\\
Assistant: The oil-rich Bakken shale formation has made North Dakota the second-largest crude-producing state behind only Texas.\\
User: Why\\
Assistant: North Dakota generated 852 million barrels of Bakken crude, The Bakken and the Three Forks formation are 94\% of Dakota's current oil production, with about 1 million barrels a day.\\
User: In how many U.S  states do we find the production\\
Assistant: Crude oil is produced in 32 U.S. states and in U.S. coastal waters.\\
User: Which ones produce the most\\
\\
\\
Question: Where is henry cavill from? \\
Dialog:\\
User: where is superman in the justice league movie\\
Assistant: In the Justice League Movie, Superman has been dead for two years and mankind is in mourning him.\\
User: who plays the character\\
Assistant: Henry Cavill plays Superman in the Justice League movie\\
User: where is the actor from\\

\bottomrule
    \end{tabular}
\end{table}

\subsection{Human dialog Generation for Bamboogle}
\label{sec:bamboogle}
Bamboogle human constructed dialogs are available here: \href{https://drive.google.com/file/d/1oUxvJ8bYL5fjKg5m7ZKJ9Wn_z0NDXQ6W/view?usp=share_link}{Bamboogle dialogs}. 

Instructions for annotator: ``Your task is to write a dialog given a question. You may use the internet in order to add factual information to dialog. The dialog intent should be understandable yet challenging. The responses should be `assistant like' -- not trying to mimic human emotions.''

\subsection{Human Evaluation Instructions}
\label{sec:human_evaluation_instructions}
Instructions for the intent-preserving task: ``You will receive a dialog and a query, and need to label 1 if the dialog has the same intent as the query. If you are unsure, you can check to see whether both intents return the same Google search results.''.

Instructions for the natural vs. auto-generated task: ``Can you guess which of the following dialogs is natural and which is auto-generated by a machine? There are 200 dialogs. 100 are synthetic, and 100 are natural. Enter 1 for Synthetic and 0 for Natural in the "synthetic?" field.''

\subsection{Additional Examples}
\label{sec:additional_examples}
\begin{table}[!htbp]
\begin{tabular}{@{}ll@{}}
\toprule
Question                                                            & BlenderBot3 Predictions                 \\ \midrule
What is the maximum airspeed (in km/h) of   the third fastest bird? & Fastest bird                      \\
Who directed the highest grossing film?                             & Top grossing films                \\
The husband of Lady Godiva was Earl of   which Anglic kingdom?      & Lady Godiva                       \\
Who founded the city where the founder of   geometry lived?         & Geometry                          \\
Who was the head of NASA during Apollo   11?                        & Apollo 11                         \\
When was the longest bridge in the world   opened?                  & Longest bridge                    \\
When was the anime studio that made Sword   Art Online founded?     & Sword Art Online                  \\
What is the capital of the country where   yoga originated?         & Yoga origin                       \\
Who is the father of the father of George   Washington?             & George Washington father          \\
Who was the first king of the longest   Chinese dynasty?            & first king of the longest dynasty \\ \bottomrule
\end{tabular}
\caption{Examples for the ground-truth queries from the Bamboogle dialogues with the BlenderBot3 queries. The BlenderBot3 seems to be trained more on finding the topic than asking concrete questions.}
\label{tab:bb3_examples}
\end{table}

Table~\ref{tab:bb3_examples} shows examples of the ground-truth vs. the predicted queries from samples in the Bamboogle dialogs dataset.

\subsection{Parlai Models}
\label{sec:parlai_models}
We used available versions in Parlai \cite{miller2017parlai} that can be accessed through an open API for a popular search engine. 
For the WizInt model we used: zoo:sea/bart\_fid\_sqse/model.
For BlendeBot3 model we used zoo:bb3/bb3\_3B/model -o gen/r2c2\_bb3.
The search engine is: \url{https://github.com/scy6500/ParlAI_SearchEngine}

\end{document}